\documentclass[10pt]{article} 
\usepackage[preprint]{tmlr}

\usepackage{amsmath,amsfonts,bm}

\def\eqref#1{equation~\ref{#1}}

\def\1{\bm{1}}

\DeclareMathAlphabet{\mathsfit}{\encodingdefault}{\sfdefault}{m}{sl}
\SetMathAlphabet{\mathsfit}{bold}{\encodingdefault}{\sfdefault}{bx}{n}

\usepackage{hyperref}
\usepackage{url}
\usepackage{secdot}
\usepackage{booktabs}
\usepackage{xspace}
\usepackage{arydshln}
\usepackage{graphicx}
\usepackage{cleveref}
\usepackage{wrapfig}
\usepackage{enumitem}

\definecolor{c1}{HTML}{e31a1c} 
\definecolor{c1a}{HTML}{fb9a99} 
\definecolor{c2}{HTML}{1f78b4} 
\definecolor{c2a}{HTML}{a6cee3} 
\definecolor{c3}{HTML}{33a02c} 
\definecolor{c3a}{HTML}{b2df8a} 
\definecolor{c4}{HTML}{6a3d9a} 
\definecolor{c4a}{HTML}{cab2d6} 
\definecolor{c5}{HTML}{ff7f00} 
\definecolor{c5a}{HTML}{fdbf6f} 
\definecolor{c9}{HTML}{ffff99} 
\definecolor{c7}{HTML}{b15928} 
\definecolor{c8}{HTML}{f781bf} 
\definecolor{c6}{HTML}{999999} 

\usepackage{fixme}
\fxsetup{
  status=draft,
  theme=color,
  silent,
}
\fxsetup{marginface=\linespread{1}\footnotesize}
\FXRegisterAuthor{bl}{ble}{\colorbox{c3a!50}{\color{c3}BL}}

\title{Does your data spark joy? Performance gains from domain upsampling at the end of training}

\author{{\sffamily Cody Blakeney\thanks{Equal contribution.  Correspondance to \{\texttt{cody.blakeney}, \texttt{mansheej.paul}, \texttt{brett.larsen}\}\texttt{@databricks.com}}\:\,, Mansheej Paul$^*$, Brett W. Larsen$^*$, Sean Owen, and Jonathan Frankle} \\
\\
{\normalfont Databricks Mosaic Research }}

\def\openreview{\url{https://openreview.net/forum?id=XXXX}} 

\begin{document}

\maketitle

\begin{abstract}
Pretraining datasets for large language models (LLMs) have grown to trillions of tokens composed of large amounts of CommonCrawl (CC) web scrape along with smaller, domain-specific datasets.
It is expensive to understand the impact of these domain-specific datasets on model capabilities as training at large FLOP scales is required to reveal significant changes to difficult and emergent benchmarks.
Given the increasing cost of experimenting with pretraining data, how does one determine the optimal balance between the diversity in general web scrapes and the information density of domain specific data?
In this work, we show how to leverage the smaller domain specific datasets by upsampling them relative to CC at the end of training to drive performance improvements on difficult benchmarks.
This simple technique allows us to improve up to 6.90 pp on MMLU, 8.26 pp on GSM8K, and 6.17 pp on HumanEval relative to the base data mix for a 7B model trained for 1 trillion (T) tokens, thus rivaling Llama-2 (7B)---a model trained for twice as long.
We experiment with ablating the duration of domain upsampling from 5\% to 30\% of training and find that 10\% to 20\% percent is optimal for navigating the tradeoff between general language modeling capabilities and targeted benchmarks.
We also use domain upsampling to characterize at scale the utility of individual datasets for improving various benchmarks by removing them during this final phase of training. 
This tool opens up the ability to experiment with the impact of different pretraining datasets at scale, but at an order of magnitude lower cost compared to full pretraining runs. 
\end{abstract}

\thispagestyle{fancy}
\fancyhead{}
\rhead{\includegraphics[width=3.75cm]{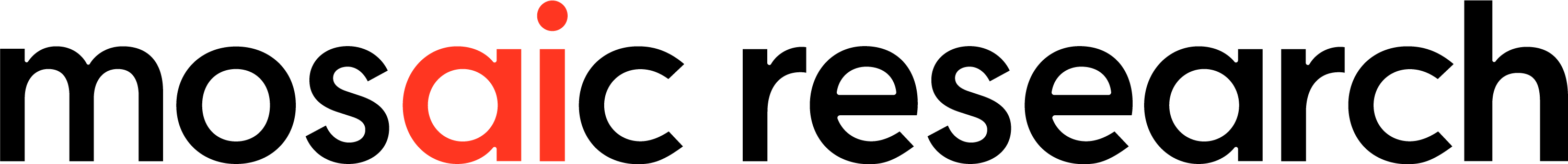}}
\lhead{\includegraphics[width=3cm]{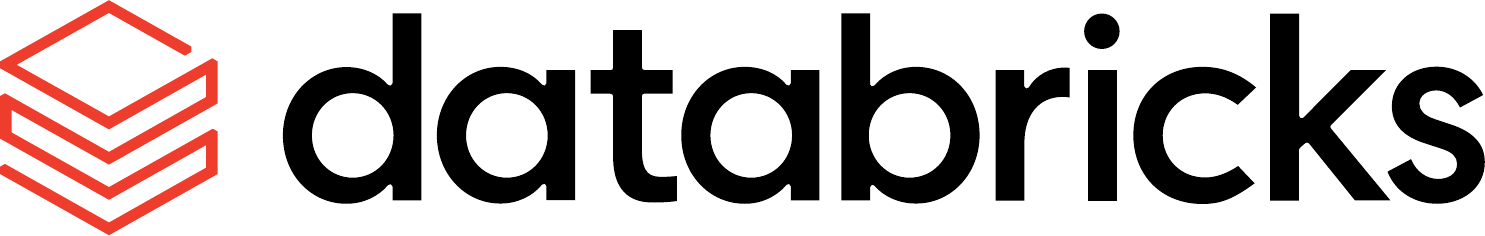}}

\section{Introduction}

Pretraining datasets for large language models (LLMs), such as Dolma \citep{soldaini2023dolma}, have grown to trillions of tokens.
To accommodate such large scales, they are typically composed of two types of data sources. 
First, they contain large amounts of web scraped data processed from CommonCrawl (CC) dumps. These are typically hundreds of billions to trillions of tokens in size and contain a diverse distribution of information. However, because of their size, they are necessarily less information dense and are not as filtered.
Second, LLM pretraining mixes contain datasets that either target certain domains or come from single high quality sources. 
These are much smaller (often less than a hundred billion tokens tokens).
They are also more carefully processed and are dense with information from domains we want LLMs to be good at; 
however, since their sources are limited, they are often less diverse \citep{together2023redpajama}.

\paragraph{Related Works:} One of the biggest challenges to pretraining LLMs is determining the optimal strategy for mixing datasets that come from CC and smaller domain specific sources.
Some previous works have opted to pretrain entirely on heavily processed CC data \citep{penedo2023refinedweb}.
Others have used different heuristics to balance between CC and more domain specific datasets \citep{together2023redpajama}.
However, most recent language models trained at scale disclose limited information on the contents of their pretraining data \citep{touvron2023llama, jiang2023mistral, jiang2024mixtral,team2024gemma}.
At smaller scales, there have been attempts to algorithmically optimize the data mix proportions, but these methods have not been openly validated at the scale most modern language models are trained \citep{xie2024doremi}.
Given the sheer cost of validating data mixing strategies at this scale, there is a paucity of open research on pretraining data for LLMs.

Ideally one would conduct data mix experiments at smaller scales to identify what is a good data mix.
However, this is often ineffective because large FLOP scales are required to reveal significant changes in difficult and emergent benchmarks.
In fact, most LLMs trained at smaller scales register random accuracy on many important benchmarks such as MMLU \citep{wei2022emergent}.
As a result, experiments at smaller scales can often be misleading; the variation between different data mixes on important benchmarks is often due to noise rather then dataset quality at this scale.
On the other hand, it is prohibitively expensive and impractical to exhaustively characterize datasets by doing multiple training runs at the scale needed to measure above random performance on these metrics.

In this work, our goal is to characterize the utility of an alternative approach to conduct pretraining data experiments at a reasonable scale.
Our strategy is to modify the data mixture \textit{at the end of training} after we have already trained for enough FLOPs to measure meaningful signal on difficult benchmarks.
We show that this is an effective strategy for improving LLM pretraining data mixes with experiments that are an order of magnitude cheaper than full training runs.

\paragraph{Contributions:} 
\begin{itemize}
\itemsep0em
    \item We begin with a baseline mix of publicly available datasets that achieves the same scaling of performance with FLOPs as the Llama-2 model family for a 7B model trained for 1 trillion tokens.
    \item We introduce domain upsampling---a data intervention which upsamples domain specific datasets relative to Common Crawl at the end of training---and demonstrate that it can boost challenging metrics. In particular, we observe improvements of up to 6.90 pp on MMLU, 8.26 pp on GSM8K, and 6.17 pp on HumanEval relative to the base data mix in our training setup.  This makes our performance comparable to Llama-2 (7B) but at approximately half the training FLOPs.
    \item We ablate the percentage of training that utilizes domain upsampling and show  10\%-20\% is optimal for navigating the tradeoff between general language modeling capabilities and targeted benchmarks.
    \item We show how domain upsampling can be used as a FLOP-efficient tool to characterize how individual datasets impact model capabilities. 
    By removing a subset of math-heavy pretraining data from the datasets we upsampled at the end of training, we quantified the impact these datasets have on specific benchmarks.
\end{itemize}

\section{Training Details}

\begin{wraptable}{r}{8.5cm}
    \small
    \centering
    \begin{tabular}{l c} \toprule 
         \textbf{Parameter} & \textbf{Value}  \\
         \midrule
         Optimizer & LionW \citep{chen2024symbolic} \\
         Learning Rate & 0.00012 \\
         Betas & 0.9, 0.95 \\
         Weight Decay & 0.00012 \\
         Max Sequence Length & 4096 \\
         Batch Size & 960 \\
         Tokenizer & Tiktoken (GPT-4) \\
         Positional Embedding & ALiBi \citep{press2022train} \\
         \bottomrule
    \end{tabular}
    \caption{Training Hyperparameters.}
    \label{tab:architecture}
\end{wraptable}

We studied domain upsampling on 7 billion parameter models trained for 1 trillion tokens.  This FLOP scale was chosen so that the model performed above the noise floor on key metrics like MMLU enabling us to see the effects of data interventions on the model. 

The 7B models trained for this work are decoder-only transformers using the MPT architecture in LLM Foundry \citep{mosaicml2023mpt}. To evaluate our models we use the latest version of the Eval Gauntlet v0.3.\citep{mosaicml2023mpt}, an evaluation framework consisting of 35 popular in context learning evaluation tasks used to evaluate LLM base models. The Gauntlet v0.3 aggregates scores on benchmarks across 6 categories. It is described in \Cref{sec:gauntlet}. We use an inverse square root learning schedule similar to \citep{zhai2022scaling}.

\section{Results}
Here we present the experiments demonstrating the performance boost achieved by domain upsampling as well as its utility in characterizing how datasets affect challenging, emergent metrics.

\subsection{Baseline data mix achieves Llama-2 scaling}
\label{sec:data-mix}

To construct a baseline data mix, we grouped a set of publicly-available datasets into 4 broad categories:
\vspace{-1em}
\begin{itemize}
\itemsep0em 
\item \textbf{Large-Scale Common Crawl:} Datasets derived from Common Crawl that emphasize scale. These datasets trade off thorough quality filtering in favor of curating a large and diverse set of tokens.
\item \textbf{Small-Scale Common Crawl:} Datasets derived from Common Crawl with more extensive filtering but are smaller than large-scale Common Crawl.
\item \textbf{Domain Specific data:} Small datasets that target certain domains or are from individual sources and are of high quality (e.g. Wikipedia).
\item \textbf{Code:} Code data across a variety of programming languages.
\end{itemize}
\vspace{-1em}

We set the proportions for mixing these datasets based on a rough heuristic for the number of epochs each of these groups would be seen during the 1 trillion token training duration.
Specifically, we choose 0.5 epochs for the Small-Scale Common Crawl and Domain Specific data and 1 epoch for Code.
The remainder of the 1 trillion tokens are filled with Large-Scale Common Crawl.
The exact proportions are in \Cref{tab:pretrain-data-proportions}.

The rationale behind choosing these proportions is as follows: we expect the Small-Scale Common Crawl and Domain Specific data to be of high quality and we wanted them to be well represented on our 1 trillion token budget. 
Also, we wanted to emphasize coding ability and so we decided to sample code data at a high percentage---initial experiments indicated that a high percentage of code around 20\% boosted programming and reasoning ability without negatively impacting language abilities.
We then treat the Large-Scale CC as filler tokens that increase the diversity of our dataset and allow us to fill our token budget.

\begin{figure}
\centering
\includegraphics[width=\linewidth]{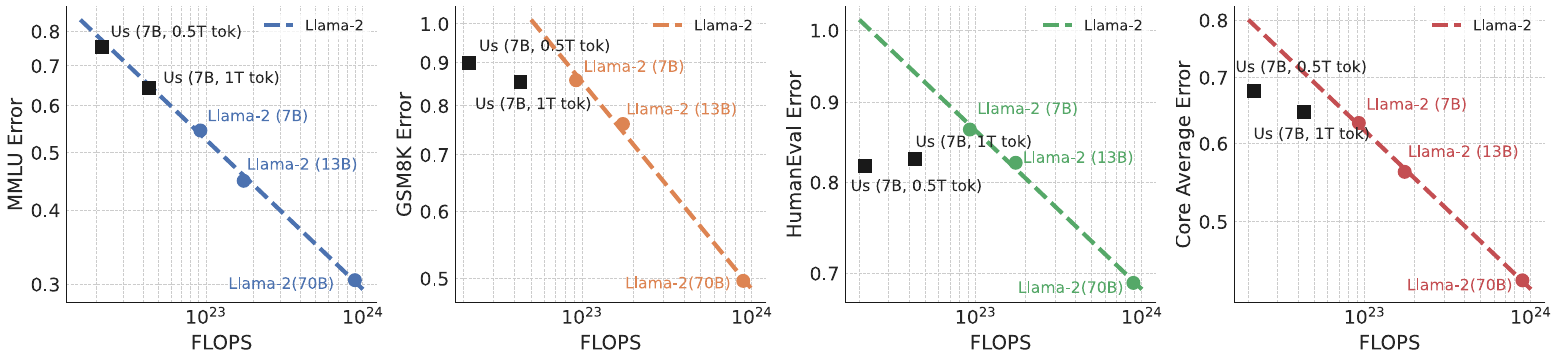}
\caption{On key benchmarks, our 7B models trained with the data mix presented in \Cref{tab:pretrain-data-proportions} have errors \emph{at or below} the error vs. FLOP scaling line of the Llama-2 family of models.  We first evaluated the performance of the 7B, 13B, and 70B variants of the Llama-2 models on MMLU, GSM8K, HumanEval, and the Gauntlet v0.3 Core Average.  We then performed linear regression on the log of the error on these metrics vs. the log of the FLOPS used to train these models.  This scaling relationship is plotted as a dashed line in the log-log plots shown above; one can observe that the models in the Llama-2 family lie close to this scaling line.  For MMLU, our models (square markers) lie on the Llama-2 scaling line.  For the other metrics, our models are significantly below the scaling line.} 
\label{fig:scaling-plots}
\end{figure}

Importantly, since the goal of our experimental setup is to demonstrate the utility of domain upsampling at the end of training (discussed in \cref{sec:du}), we opt for choosing a reasonable heuristic for picking our initial data mix proportions without too much optimization.
\Cref{tab:data-mix-eval} and \Cref{fig:scaling-plots} show the performance of this initial pretraining data mix for two 7B models trained for 0.5T and 1T tokens.
This heuristic has indeed been validated by our empirical results; 
plotting error vs. FLOPs shows that our models lie on or below the Llama-2 scaling line on the Gauntlet v0.3 Core Average, MMLU, GSM8K, and HumanEval.
Interestingly, though the overall performance scaling (as measured by Gaunlet v0.3 Core Average) is very similar, our particular data choices and mixing coefficients have led to slightly different tradeoffs.
The model trained for 1T tokens outperforms the Llama-2 7B model trained for 2T tokens on GSM8K and HumanEval. This indicates that our models have better mathematical and programming ability despite being trained for half the number of tokens.
We also provide a comparison to OpenLlama 7Bv2 \citep{openlm2023openllama}, a 7B model that provides some open details about their data mix.

\begin{table}
    \centering
    \small
    \begin{tabular}{l c c c} \toprule 
         \textbf{Dataset Category} & \textbf{Percentage} & \textbf{Tokens} & \textbf{Epochs (1T)}  \\ 
         \midrule
         \textit{Large-Scale Common Crawl} & 34.35\% & 343.5B & 0.148  \\
         \textit{Small-Scale Common Crawl} & 36.70\% & 367.0B & 0.5  \\
         \textit{Domain Specific} & 7.17\% & 71.7B & 0.5  \\
         \textit{Code} & 21.78\% & 217.8B & 1 \\
         \bottomrule
    \end{tabular}
    \caption{Proportions for our pretraining data mix in terms of the 4 dataset groups. Code data was included at twice the proportion of other domain specific datasets to focus on boosting coding capabilities.  Large-scale Common Crawl was used to fill the remainder of the tokens once the other proportions were chosen.}
    \label{tab:pretrain-data-proportions}
\end{table}

\begin{table}[!htp]
    \centering
    \small
    \begin{tabular}{l c c c c} \toprule 
         &  \multicolumn{2}{c}{\textbf{Us} (7B)} & \textbf{Llama-2} & \textbf{OpenLlama}  \\
         \cmidrule(lr){2-3}
         \cmidrule(lr){4-4}
         \cmidrule(lr){5-5}
         \textbf{Benchmark}  & 0.5T tok & 1T tok & 7B (2T tok) &  7Bv2 (1T tok)   \\ \midrule 
         MMLU (5-shot)  & 24.70 & 35.69 & {\bf \color{c2} 45.51} & 40.38   \\
         GSM8K (8-shot) & 10.16 & {\bf \color{c2} 14.71} & 14.25 & 7.05   \\
         HumanEval (pass@1) & {\bf \color{c2} 18.02} & 17.23 & 13.55 & 15.20  \\
         \midrule
         \textit{Gauntlet v0.3} \\
         \quad Core Average & 32.13 & 35.37 & {\bf \color{c2} 37.05} & 32.96  \\
         \quad World Knowledge  &  39.29 & 41.77 &  {\bf \color{c2} 50.94} & 43.79   \\
         \quad Commonsense Reasoning  &  30.52 & {\bf \color{c2} 38.38} & 35.48  & 34.91 \\
         \quad Language Understanding & 61.47  & 61.52 &  {\bf \color{c2} 65.02} & 61.00  \\
         \quad Symbolic Problem Solving & 14.10 & 16.28 &  {\bf \color{c2} 22.23} & 19.09   \\
         \quad Reading Comprehension   & 29.36 & {\bf \color{c2} 37.02} & 35.05 & 23.82   \\
         \quad Programming (HE) & {\bf \color{c2} 18.02} & 17.23 & 13.55 & 15.20 \\
         \bottomrule
    \end{tabular}
    \caption{Full evaluation results for the models presented in \Cref{fig:scaling-plots}.  We note that a 7B model trained with our data mix for 1T tokens outperforms Llama2-7B---a model trained for 2T tokens---on GSM8K, HumanEval, and the Commonsense Reasoning and Reading Comprehension subsets of the Gauntlet v0.3. We also compare to OpenLlama 7Bv2, a similar model with a publicly available data mix trained for 1T tokens.  Note that HumanEval (HE) is the sole component of the programming section of the Gauntlet.}
    \label{tab:data-mix-eval}
\end{table}

\subsection{Domain upsampling significantly boosts performance on challenging metrics}
\label{sec:du}

\begin{wraptable}{r}{8.5cm}
    \small
    \centering
    \begin{tabular}{l c c c c} \toprule 
         \textbf{Dataset Category} & \textbf{Percentage} & \textbf{Tokens}  \\ 
         \midrule
         \textit{Large-Scale Common Crawl} & 0\% & 0  \\
         \textit{Small-Scale Common Crawl} & 30\% & 60B \\
         \textit{Domain Specific} & 35\% & 70B  \\
         \textit{Code} & 35\% & 70B \\ 
         \bottomrule
    \end{tabular}
    \caption{Domain upsampling (DU) is a data intervention in which datasets are removed from the data mix at the end of training in order to scale up or upsample the reamining data.  We consider results for removing Large-Scale Common Crawl and scaling up the remaining datasets as specified in this table.}
    \label{tab:du-data-proportions}
\end{wraptable}

Next, we introduce domain upsampling during the last 20\% of training for our 1T token training run. 
For this, we start with a checkpoint at 0.8T tokens of training, change the mixing proportions of our pretraining data mix, and continue training for the remaining 0.2T tokens.
The exact mixing proportions of our domain upsampled pretraining mix are in \Cref{tab:du-data-proportions}.
These percentages were chosen based on the following heuristic: we hypothesize that though the Large-Scale CC adds a lot of diversity to the pretraining data mix, it is advantageous to emphasize Domain Specific data at the end of training to bias our model towards token distributions that have high information density in domains we care about.
Thus, we remove Large-Scale Common Crawl from our data mix while upsampling both Domain Specific and Code subsets. 
We also maintain Small-Scale Common Crawl at high percentage to prevent a large distribution shift in our pretraining data.

\begin{figure}
\centering
\includegraphics[width=\linewidth]{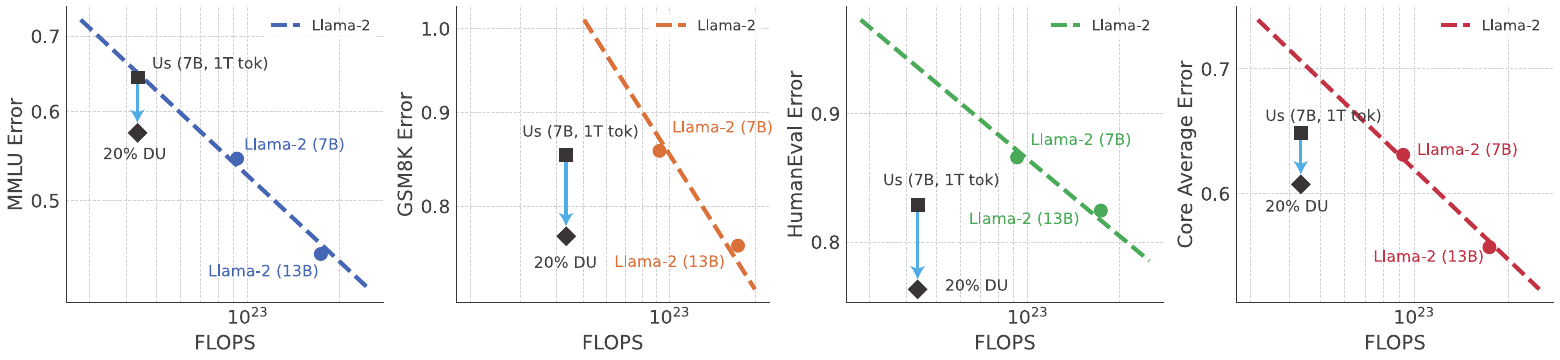}
\caption{Domain upsampling (DU) using the proportions presented in \Cref{tab:du-data-proportions} provides a significant performance boost on challenging metrics for no additional FLOP cost.  The dashed lines represent the same scaling for the Llama-2 family of models as described in \Cref{fig:scaling-plots}.  The square markers are the performance of our 7B model trained for 1T tokens with the data mix described in \Cref{sec:data-mix}; the diamond markers are the resulting models when domain upsampling is performed with the proportions specified in \Cref{tab:du-data-proportions} for the final 20\% or 200B tokens of training.  The light blue arrow emphasizes the improvement we observe from DU: \textbf{6.90 pp} on MMLU, \textbf{8.26 pp} on GSM8K, \textbf{6.17 pp} on HumanEval, and \textbf{3.95 pp} on the Gauntlet v0.3} 
\label{fig:scaling-plots-2}
\end{figure}

\begin{table}[!htp]
    \centering
    \begin{tabular}{l c c c c} \toprule 
         &  \multicolumn{2}{c}{\textbf{Us} (7B, 1T tok)} & \textbf{Llama-2} & \textbf{OpenLlama} \\
         \cmidrule(lr){2-3}
         \cmidrule(lr){4-4}
         \cmidrule(lr){5-5}
         \textbf{Benchmark}  & No DU & 20\% DU & 7B (2T tok) & 7Bv2 (1T tok)   \\ \midrule 
         MMLU (5-shot)  & 35.69 & 42.59 & {\bf \color{c2} 45.51} & 40.38 \\
         GSM8K (8-shot) & 14.71 & {\bf \color{c2} 22.97} & 14.25 & 7.05 \\
         HumanEval (pass@1) & 17.23 & {\bf \color{c2} 23.40} & 13.55 & 15.20\\
         \midrule
         \textit{Gauntlet v0.3} \\
         \quad Core Average & 35.37 & {\bf \color{c2} 39.32} & 37.05 & 32.96   \\
         \quad World Knowledge  & 41.77 & 44.19 & {\bf \color{c2} 50.94} & 43.79  \\
         \quad Commonsense Reasoning    & 38.38 & {\bf \color{c2} 42.59} & 35.48 & 34.91  \\
         \quad Language Understanding   & 61.52 & 60.08 & {\bf \color{c2} 65.02} & 61.00 \\
         \quad Symbolic Problem Solving & 16.28 & 20.23 & {\bf \color{c2} 22.23} & 19.09  \\
         \quad Reading Comprehension    & 37.02 & {\bf \color{c2} 45.45} & 35.05 & 23.82  \\
         \quad Programming (HE) & 17.23 & {\bf \color{c2} 23.40} & 13.55 & 15.20\\
         \bottomrule
    \end{tabular}
    \caption{Full evaluation reults for the models presented in \Cref{fig:scaling-plots-2} along with a comparison to OpenLlama 7Bv2.  Overall, our model with 20\% domain upsampling outperforms Llama2 (7B) on the Gauntlet v0.3 despite being trained for 1T fewer tokens.  Our model particularly excels at GSM8K and HumanEval but still trails Llama-2 (7B) on MMLU.}
    \label{tab:cl-model-eval2}
\end{table}

The results of this end-of-training data intervention are shown in \Cref{tab:cl-model-eval2} and \Cref{fig:scaling-plots-2}. 
Domain upsampling was incredibly effective in boosting model performance relative to the initial pretraining data mix on all challenging benchmarks.
Given the large amount of code and math related data in the domain upsampled data mix, it is perhaps unsurprising that this intervention led to GSM8K and HumanEval scores that are approximately 10pp higher than Llama-2 (7B) despite the model being trained for half the total number of tokens.
Additionally, this did not come at a cost to general language modeling capabilities; it led to an overall model performance improvement as measured by Gauntlet v0.3 Core Average. 
In fact, it improved world knowledge---as measured by MMLU and the Gauntlet v0.3 subset---relative to the base data mix, bringing us closer to Llama-2 (7B) performance on these metrics.
There was only a small 1pp tradeoff in the Language Understanding subset.

Overall, this across the board improvement on challenging benchmarks establishes the efficacy of domain upsampling as a pretraining data intervention for improving model performance.
Importantly, even using simple heuristics for choosing the new data mix proportions has strong positive effects, leaving opportunity for further improvement with better tuned mixing proportions.

\subsection{Changing the duration of domain upsampling enables us to navigate the trade-off between targeting specific domains and general purpose language models}

\begin{figure}[!h]
\centering
\includegraphics[width=\linewidth]{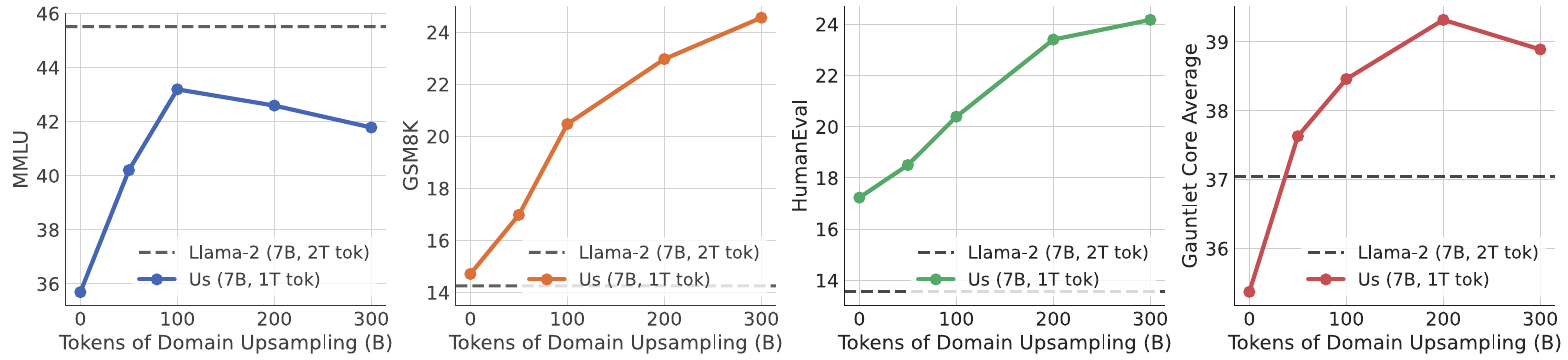}
\caption{Ablating the duration of Domain Upsampling (DU).  Here we consider performing DU for the final 5\%, 10\%, 20\%, and 30\% (50B, 100B, 200B, and 300B tokens) of training for a 7B model for a total duration 1T tokens.  We observe that GSM8K and HumanEval performance continue to improve with increased DU while MMLU and the Gauntlet Core Average peak at 10\% and 20\% respectively.  Looking across the metrics presented in \Cref{tab:cl-model-eval2}, we conclude that (1) DU for the final 10\%-20\% of training provides the best trade-off for this set up and (2) the mix used for DU should not be used for the entire duration of training.} 
\label{fig:domain-upsampling-ablation}
\end{figure}

The success of domain upsampling for the last 20\% of training raises the question: are the improvements from an end-of-training data intervention or are they from overall better data mix proportions?
Phrased another way, are the data mix proportions in \Cref{tab:du-data-proportions} better than our initial data mix and would training a model for 1T tokens with this data mix lead to better performance?
In this section, we provide evidence that this is not the case and in fact, treating domain upsampling as an end-of-training data intervention helps us better tradeoff domain specific improvements and general language modeling capabilities.

\begin{table}[!h]
    \centering
    \begin{tabular}{l c c c c c} \toprule 
         &  \multicolumn{5}{c}{\textbf{Us} (7B, 1T tok)} \\
         \cmidrule(lr){2-6}
         \textbf{Benchmark}  & 0\% &  5\% & 10\% & 20\% & 30\%  \\ \midrule 
         MMLU (5-shot)  & 35.69 & 40.20 & {\bf \color{c2} 43.19} & 42.59 & 41.78  \\
         GSM8K (8-shot) & 14.71 & 16.98 & 20.47 & 22.97 & {\bf \color{c2} 24.56}  \\
         HumanEval (pass@1) & 17.23 & 18.50 & 20.39 & 23.40 & {\bf \color{c2} 24.17} \\
         \midrule
         \textit{Gauntlet v0.3} \\
         \quad Core Average & 35.37 & 37.63 & 38.46 & {\bf \color{c2} 39.32} & 38.89             \\
         \quad World Knowledge          & 41.77 & 43.52 & {\bf \color{c2} 44.72} & 44.19 & 43.71  \\
         \quad Commonsense Reasoning    & 38.38 & 42.97 & 42.33 & {\bf \color{c2} 42.59} & 42.19   \\
         \quad Language Understanding   & {\bf \color{c2} 61.52} & 61.05 & 60.41 & 60.08 & 60.35  \\
         \quad Symbolic Problem Solving & 16.28 & 18.50 & 19.55 & 20.23 & {\bf \color{c2} 20.44}  \\
         \quad Reading Comprehension    & 37.02 & 41.23 & 43.35 & {\bf \color{c2} 45.45} & 42.50  \\
         \quad Programming (HE) & 17.23 & 18.50 & 20.39 & 23.40 & {\bf \color{c2} 24.17} \\
         \bottomrule
    \end{tabular}
    \caption{Full evaluation results for the models presented in \Cref{fig:domain-upsampling-ablation}.}
    \label{tab:cl-model-eval-dur}
\end{table}

To identify when in training this intervention should be applied, we ablate our previous experiment by performing domain upsampling for the last 5\%, 10\%, 20\%, and 30\% of training. 
The results of this experiment are shown in \Cref{fig:domain-upsampling-ablation} and \Cref{tab:cl-model-eval-dur}.
Note, while the math and programming related benchmarks, such as HumanEval, GSM8K and related Gauntlet v0.3 subscores, continue to improve as we increase the fraction of training that uses domain upsampling, other benchmarks reach optimal performance at 20\% or less.
For example, MMLU peaks at 10\% and Gauntlet v0.3 Core Average peaks at 20\%.
Thus, as we increase the fraction of training with domain upsampling beyond 20\%, improvements on math and coding benchmarks come at the cost of performance on general language modeling abilities.

This apparent trade-off indicates that the domain upsampling data mix proportions are not incontrovertibly better than the initial data mix, and training with it for the full 1T token duration would not lead to a better general purpose language model.
We do not rule out that there is an alternate mix that achieves similar performance as the 20\% domain upsampling experiment when trained for the full training duration.
However, finding such a mix is expensive to iterate on for the full training run. 
Thus, the strength of domain upsampling is that it gives us a tool to navigate this tradeoff between targeted domains and general language modeling abilities with experiments that are an order of magnitude cheaper.

\subsection{Domain upsampling is a FLOP-efficient tool to characterize how individual datasets impact model capabilities}
\label{sec:sans-math}

Having observed that upsampling code and our domain specific datasets for a small percentage of training leads to significant improvements on difficult and emergent tasks, we explore the question: how does one attribute improvements to specific subsets of these data? Notably, as can be seen in \Cref{fig:domain-upsampling-ablation} and \Cref{tab:cl-model-eval2}, GSM8K scores---a task measuring math and reasoning abilities---improves monotonically as duration of domain upsampling is increased. We hypothesize, given the quantity of math related data in our high-quality datasets that these may be responsible for some or all of this improvement. To quantify the impact of these datasets we repeat our experiment, applying domain upsampling for the last 10\% of the training duration. We keep our dataset proportions identical to those in \Cref{tab:du-data-proportions}, but remove the math related subsets. We present the results in \Cref{tab:cl-model-eval3}.

\begin{table}[!h]
    \centering
    \begin{tabular}{l c c c} \toprule 
         & &  \multicolumn{2}{c}{10\% DU} \\
         \cmidrule(lr){3-4}
         \textbf{Benchmark}  & No DU & With Math & Sans Math   \\ \midrule 
         MMLU (5-shot)  & 35.69 & {\bf \color{c2} 43.19} & 29.71 \\
         GSM8K (8-shot) & 14.71 & 20.47 & 11.37  \\
         HumanEval (pass@1) & 17.23 & 20.39 & {\bf \color{c2} 21.15} \\
         \midrule
         \textit{Gauntlet v0.3} \\
         \quad Core Average &  35.37 & {\bf \color{c2} 38.46} & 32.54            \\
         \quad World Knowledge          & 41.77 & {\bf \color{c2} 44.72} & 39.08 \\
         \quad Commonsense Reasoning    & 38.38 & {\bf \color{c2} 42.33} & 31.76   \\
         \quad Language Understanding   & {\bf \color{c2} 61.52} & 60.41 & 59.97  \\
         \quad Symbolic Problem Solving & 16.28 & {\bf \color{c2} 19.55} & 16.80  \\
         \quad Reading Comprehension    & 37.02 & {\bf \color{c2} 43.35} & 26.48   \\
         \quad Programming & 17.23 & 20.39 & {\bf \color{c2} 21.15} \\
         \bottomrule
    \end{tabular}
    \caption{Removing the math-specific datasets during domain upsampling results in significantly worse performance on all metrics except programming vs. performing domain upsampling with these datasets.  Experiments such as this provide a significantly cheaper method to characterize datasets compared to full pretraining runs with different data mixes.  Furthermore, unlike chepaer experiment with smaller models, we get signal on the effects of the datasets on challenging benchmarks like MMLU, GSM8K and HumanEval.}
    \label{tab:cl-model-eval3}
\end{table}

We observe that not only do the the mathematical knowledge and reasoning skills, as measured by MMLU (which contains STEM subsets) \& GSM8k, \textit{not} reach the same level of performance as the model trained using domain upsampling that included them, but in fact performance is worse then the baseline model with no domain upsampling. Moreover, every Gauntlet v0.3 subcategory score for the domain upsampling sans-math with the exception of programming is lower than the baseline model. From this we can draw the conclusion that these specific datasets are responsible for the majority of the mathematical knowledge and reasoning capabilities in both the base model and the domain upsampled variant.

With this observation we have successfully done something which generally would be considerably more expensive. That is, we have measured the impact of pretraining datasets at a scale where difficult and emergent tasks can be reliably measured, but at an order of magnitude fewer training FLOPS. We believe application of domain upsampling opens up the ability for researchers to experiment with their pretraining datasets in a tractable way as compared to full pretraining runs.

\section{Discussion}

Pretraining LLMs has become an increasingly costly and clandestine endeavor given the scale of compute required for each experiment.
This problem is exacerbated by the multi-faceted decision space presented to practitioners, especially in the selection of pretraining data.
Since many important model capabilities emerge with scale, trying to explore this design space at small compute budgets is often ineffective: observations made about the effects of the pretraining data mix typically do not transfer to larger models or training budgets.

In this work, we consider a baseline data mix of publicly available datasets that achieves or exceeds the scaling of the Llama-2 family of models on key benchmarks.
Next, we take a crucial first step towards making experimentation with pretraining datasets cheaper.
We introduce domain upsampling, a method that can strongly impact the performance of the model by making targeted changes to the data mix at the end of training.
This enables us to achieve the performance of Llama-2 (7B) but with half the training budget.
By varying the duration of domain upsampling, we demonstrate how to navigate the tradeoff between targeting specific domains and making general purpose language models.

Finally, we show how making changes to the data mix only during the domain upsampling period enabled us to cheaply characterize the impact of several math-focused datasets, and we see many opportunities to use this method as a general tool for studying pretraining data in a FLOP-efficient manner.
It also creates a platform to test data interventions at scale: instead of testing possible dataset optimization algorithms at small scales and hoping they will generalize, we can test them at the end of training to effectively measure their impact at scale. 
By bringing down the cost of experimentation we have made pretraining data experiments more accessible, and
we will release our models and intermediate checkpoints as research artifacts to the community as a resource to unlock further insights into LLM pretraining data.

\newpage

\setcitestyle{numbers}
\bibliography{references}
\bibliographystyle{tmlr}

\newpage

\appendix

\setcitestyle{authoryear}
\section{Gauntlet v0.3}
\label{sec:gauntlet}

The Gauntlet v0.3 is a aggregation of benchmark developed by Mosaic Research. Rather than reporting a monolithic metric in which all scores are aggregated together, the individual benchmarks were grouped into six broad competencies corresponding to different capabilities we want our LLMs to have:  

\begin{enumerate}
    \item \textbf{World Knowledge:} Measures the model's factual knowledge across a range of subjects.
    \item \textbf{Commonsense Reasoning:} Evaluates the model's ability to do basic reasoning tasks that require commonsense knowledge of objects, their properties, and their behaviors.
    \item \textbf{Language Understanding:} Assesses the model's ability to understand structure and properties of language.
    \item \textbf{Symbolic Problem Solving:} Tests the model's ability to solve a diverse range of symbolic tasks including arithmetic, logical reasoning, algorithms, and algebra.
    \item \textbf{Reading Comprehension:} Measures a model's ability to answer questions based on information in a passage of text.
    \item \textbf{Programming:} Quantifies the ability to generate code from docstring descriptions. 
\end{enumerate}

These divisions allow for more fine-grained comparison between models and is especially useful for understanding how datasets affect different capabilities of the model.  The random baseline of each metric was subtracted out before aggregating. 
For example, if the metric is 4-option multiple choice questions giving a random baseline of 25\% and the model achieves 30\% then this would be aggregated as $(0.3-0.25)/(1-0.25) = 0.0667$, essentially rescaling accuracy above change to be between 0 and 1.  If the random baseline is approximately 0, then the metric is reported as is. \Cref{tab:gauntlet-v3} list the benchmarks in each category.

\begin{table}[!ht]
    \centering
    \begin{tabular}{l p{6cm}} \toprule 
         \textbf{Benchmark} & \textbf{Citation}  \\ \midrule 
         \textit{World Knowledge} \\
         \quad Jeopardy (3-shot) & \citep{wolfe22jeopardy} \\
         \quad MMLU (5-shot) & \citep{hendrycks2020measuring} \\
         \quad BIG-bench Wikidata (3-shot) & \citep{srivastava2022beyond} \\
         \quad ARC-easy (3-shot) & \citep{clark2018think} \\
         \quad ARC-challenge (3-shot) & \citep{clark2018think} \\
         \quad TriviaQA-Subsampled (3-shot) & \citep{joshi2017triviaqa} \\
         \midrule
         \textit{Commonsense Reasoning} \\
         \quad BIG-bench Strategy QA & \citep{srivastava2022beyond} \\
         \quad BIG-bench Strange Stories & \citep{srivastava2022beyond} \\
         \quad COPA (0-shot) & \citep {roemmele2011choice} \\
         \quad PIQA (10-shot) & \citep{bisk2020piqa} \\
         \quad SIQA (3-shot) & \citep{sap2019socialiqa} \\
         \quad Openbook QA (10-shot) & \citep{mihaylov2018can} \\
         \quad Commonsense QA (0-shot) &  \citep{talmor2018commonsenseqa} \\
         \midrule
         \textit{Language Understanding} \\
         \quad LAMBADA & \citep{paperno2016lambada} \\
         \quad HellaSwag & \citep{zellers2019hellaswag} \\
         \quad Winograd (3-shot) & \citep{levesque2012winograd} \\
         \quad Winogrande (5-shot) & \citep{sakaguchi2021winogrande} \\
         \midrule
         \textit{Symbolic Problem Solving} \\
         \quad BIG-bench Elementary Math QA (1-shot) & \citep{srivastava2022beyond} \\
         \quad BIG-bench Dyck Languages (5-shot) & \citep{srivastava2022beyond}\\
         \quad BIG-bench Operators (3-shot) & \citep{srivastava2022beyond} \\
         \quad Simple Arithmetic (with spaces, 5-shot) & \citep{mosaicml2023mpt} \\
         \quad Simple Arithmetic (no spaces, 5-shot) & \citep{mosaicml2023mpt}  \\
         \quad GSM8K (8-shot) & \citep{cobbe2021training} \\
         \quad SVAMP (5-shot) & \citep{patel2021svamp} \\
         \quad AGI Eval LSAT AR (5-shot) & \citep{zhong2023agieval} \\
         \midrule
         \textit{Reading Comprehension} \\
         \quad SQuAD (3-shot) & \citep{rajpurkar2016squad} \\
         \quad BoolQ & \citep{clark2019boolq} \\
         \quad CoQA & \citep{reddy2019coqa} \\
         \quad AGI Eval LSAT RC (5-shot) & \citep{zhong2023agieval} \\
         \quad AGI Eval LSAT LR (5-shot) &  \citep{zhong2023agieval} \\
         \quad AGI Eval SAT En (5-shot) & \citep{zhong2023agieval} \\
         \midrule 
         \textit{Programming} \\
         \quad HumanEval (pass@1) & \citep{chen2021evaluating} \\
         \bottomrule
    \end{tabular}
    \caption{Metrics included in the Gauntlet v0.3. Evaluation metrics are 0-shot unless otherwise denoted.}
    \label{tab:gauntlet-v3}
\end{table}


\end{document}